# Advanced Financial Fraud Detection Using GNN-CL Model

Yu Cheng[1], Junjie Guo[2], Shiqing Long[3], You Wu[4], Mengfang Sun[5], Rong Zhang[6]

[1]Columbia University, USA

[2]Rutgers University, USA

[3]Independent Researcher, China

[4]College of William & Mary, USA

[5]Stevens Institute of Technological, USA

[6]University of California, Davis, USA,

**Abstract:** The innovative GNN-CL model proposed in this paper marks a breakthrough in the field of financial fraud detection by synergistically combining the advantages of graph neural networks (gnn), convolutional neural networks (cnn) and long short-term memory (LSTM) networks. This convergence enables multifaceted analysis of complex transaction patterns, improving detection accuracy and resilience against complex fraudulent activities. A key novelty of this paper is the use of multilayer perceptrons (MLPS) to estimate node similarity, effectively filtering out neighborhood noise that can lead to false positives. This intelligent purification mechanism ensures that only the most relevant information is considered, thereby improving the model's understanding of the network structure. Feature weakening often plagues graph-based models due to the dilution of key signals. In order to further address the challenge of feature weakening, GNN-CL adopts reinforcement learning strategies. By dynamically adjusting the weights assigned to central nodes, it reinforces the importance of these influential entities to retain important clues of fraud even in less informative data. Experimental evaluations on Yelp datasets show that the results highlight the superior performance of GNN-CL compared to existing methods.

## I. INTRODUCTION

In today's digital financial era, with the vigorous development of Internet finance, mobile payment, and e-commerce, the convenience and popularity of financial services have been improved unprecedentedly. However, this series of progress also brings new challenges to financial security, especially the diversification and concealment of financial fraud, which seriously threatens the stability of the financial system and the safety of users' property. Financial institutions around the world report tens of billions of dollars in economic losses from fraudulent activities every year, and this number continues to grow. Therefore, building an efficient and accurate financial fraud detection system has become an urgent need to maintain financial security [1].

As a revolutionary technology in the field of artificial intelligence, deep learning has made remarkable achievements in many fields such as image recognition[2-4] and natural language processing[5-8] with its powerful data processing ability, automatic feature extraction, and highly abstract learning mechanism. In recent years, its application in the field of financial fraud detection has also attracted increasing attention. Analyzing a large number of transaction data through deep neural networks can capture complex patterns and abnormal behaviors that are difficult to be found by traditional methods, thereby improving the accuracy and efficiency of fraud detection.

However, the application of deep learning algorithms in financial fraud detection is not smooth. Problems such as data imbalance, feature selection, insufficient generalization ability of the model, and the continuous evolution of fraud methods pose severe challenges to the effectiveness and stability of the model. Therefore, how to optimize the deep learning algorithm to build a more accurate and robust financial fraud risk assessment model has become a research hotspot of common concern in academia and industry.

This paper aims to deeply explore the application of deep learning algorithms in financial fraud detection, focusing on the optimization of risk assessment models. This paper proposes a new risk assessment framework combining adaptive feature learning, attention mechanism and ensemble learning to solve the problem of data imbalance, enhance the learning ability of the model for complex fraud patterns, and improve the generalization performance of the model. In addition, this paper will also conduct experimental verification through actual financial transaction data sets, and compare and analyze the performance differences between the proposed model and the existing methods, in order to provide a more advanced and practical fraud detection solution for the financial industry.

## II. CORRELATIONAL RESEARCH

With the advent of the era of big data and the rapid development of machine learning technology, deep learning algorithms have shown great potential in the field of financial fraud detection because of their excellent performance in data processing ability and pattern recognition ability. In contrast to conventional machine learning techniques, deep architectures comprising multiple layers provide enhanced learning capabilities. These advanced structures have been successfully applied across various domains, including long short-term memory (LSTM) networks [9-11], computer vision [12-14], and medical diagnosis [15-16]. In view of the diversity and complexity of financial fraud, the risk assessment model needs to have higher accuracy and rapid response, and deep learning technology can meet this series of requirements.

In recent years, due to the obvious limitations of rule-based fraud detection methods, many researchers have turned to Machine Learning (ML) methods to improve the ability of fraudulent transaction detection. Through in-depth analysis of money laundering characteristics, the decision tree technology is applied to anti-money laundering practice. Ketenci et al. [17] innovatively introduced a set of feature sets based on time-frequency analysis in order to improve the efficiency of anti-money laundering system. This set uses the binary representation of financial transactions, combines random forest as a machine learning tool, and uses simulated annealing algorithm for hyperparameter tuning. This study proves the utility of time-frequency features for distinguishing suspicious and non-suspicious entities. Lee et al. [18] used supervised machine learning algorithm to classify illegal nodes in the Bitcoin network environment, and specifically used random forest and artificial neural network to identify illegal transactions. By integrating the statistical characteristics of trading volume in a fixed time window, the entropy value was formed, and the derivative characteristics of entropy were used as the model input to further enrich the identification dimension of fraud. Le et al. [19] proposed a composite method combining clustering analysis and Multi-layer Perceptron (MLP), which used a simple center-based clustering technique to screen suspected money laundering cases, and used the frequency and amount of transactions as the training features of the MLP model.

These studies highlight the broad impact of machine learning on financial fraud prevention and highlight how model design is evolving from multiple perspectives to accommodate the complexity and variability of fraud. Although traditional methods focus on the statistical attribute analysis of users and ignore the importance of the relationship between users, the development of graph neural network technology makes up for this defect. For example, the MvMoE model introduced by Liang et al. [20] integrates heterogeneous financial data through a multi-view network with hierarchical attention mechanism, which promotes the effective extraction of information and the collaborative improvement of task performance. Inspired by the connected subgraph strategy, the work of Liu et al. [21] dynamically learns feature representations from the account device network for the attacker's behavior patterns. Wang et al. used RNN to mine the temporal patterns of user click sequences to enhance the accuracy of behavior analysis [22]. Yvan Lucas[23] applied multi-view HMM to construct features automatically, which improved the efficiency of credit card fraud recognition. The champion-challenger framework proposed by Eunji Kima[24] fuses ensemble learning with deep learning models, demonstrating the advantages of cross-method cooperation in complex fraud detection tasks. The work of Van Vlasselaer[25] comprehensively considers a variety of transaction characteristics, including cardholder habits and geographical locations, and enhances the accuracy of the model through network analysis. Song Yang [26] and Ju Chunhua [27] optimized the fraud detection model from the aspects of feature selection, data balance and sequence learning with the help of SVM, k-means and LSTM combined with KNN-SMOTE technology, and achieved more refined classification and performance improvement. Together, these studies advance the frontier of fraud detection techniques in fintech, continuously exploring more comprehensive and efficient solutions.

III. METHOD

The model structure GNN-CL proposed in this paper is innovative, as shown in Figure 1. The core of the model consists of three layers of GNN, CNN and LSTM, and is designed around three key modules: neighborhood noise purifier, core node intensifier and relationship summarizer. The neighborhood noise purifier uses MLP to construct a similarity evaluation mechanism, which discriminates and reduces the noise in the neighborhood information according to the similarity between nodes. Despite the initial filtering, the impact of noise still exists, and it is urgent to optimize the strategy to mitigate its interference and ensure the balance between the core node and its neighbors. In view of this, the core node strengthening mechanism comes into being, which implements a weighting strategy for core nodes when aggregating neighborhood information, aiming to weaken the adverse effect of residual noise, ensure the brightness of core node characteristics in the fusion process, and avoid information flooding.

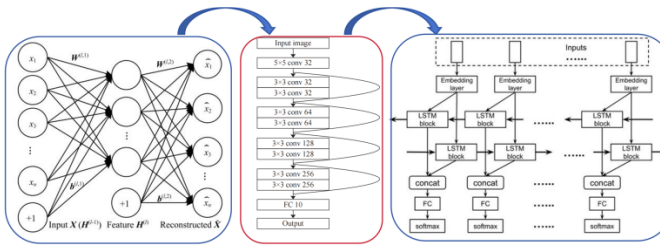

Figure 1. GNN-CL model architecture

A. GNN layer

1. Neighborhood noise purifier

Fraudulent behaviors often evade the monitoring of anti-fraud systems by changing the usual operation mode or imitating normal behaviors, such as implementing relationship disguise strategies [28]. Fraudsters can evade the screening of typical detection models by establishing dense connections with normal users. However, the neglect of such strategies in early models (e.g., GCN, GAT) undoubtedly adds difficulty to the identification of fraud.

(1) Similarity measure method

To deal with the challenge of feature and relation masquerading, many similarity measures have been proposed, including the classic cosine similarity. However, the limitation of cosine similarity is that it only focuses on the direction of the vector rather than its magnitude, fails to fully consider the scale change of different user feature values, and ignores the potential value of node label information in similarity calculation. In view of this, this paper introduces an improved similarity measurement strategy, which uses Multi-layer Perceptron (MLP) in each layer as the prediction model, and uses the L1 distance between the prediction results to quantify the similarity between nodes, aiming to evaluate the relationship between nodes more comprehensively, overcome the limitations of the original method, and improve the accuracy and robustness of fraud detection.

$$D^l(v,u) = \left\| L^l\left(h_v^{(l-1)}\right) - L^l\left(h_u^{(l-1)}\right) \right\|_1 \quad (1)$$

In the formula, $L^l(h_v^{(l-1)}) = \sigma(MLP^l((h_v^{(l-1)})))$ represents the similarity score of the feature embeddings obtained by node v at the l-th layer after being transformed by the multi-layer perceptron MLP. Here, σ stands for the activation function. However, a large value of $D^l(v,u)$ indicates that the similarity between node v and node u is weak. To enhance the intuition, we can simplify the similarity measure statement as follows.

$$Similarity_l(v,u) = \frac{1}{1 + D_l(v,u)} \quad (2)$$

Therefore, when $Similarity_l(v,u)$ is close to 1, it means that node v and node u are highly similar. If $Similarity_l(v,u)$ is close to 0, it means that the two nodes are significantly different, which intuitively shows the closeness of the association between nodes. In view of the fact that we use a learnable MLP to estimate the similarity score, in order to optimize this process, we introduce the cross-entropy loss function as a key indicator to measure the performance of the neighbor noise filter, so as to guide the model to better distinguish and reduce the interference of irrelevant neighbors during training.

$$\mathcal{L}_D^{(l)} = -\sum_r^R \sum_{v \in V} y_v log\left(MLP^{(l)}\left(h_{v,r}^{(l)}\right)\right) + (1 - y_v) log\left(1 - \left(MLP^{(l)}\left(h_{v,r}^{(l)}\right)\right)\right) \quad (3)$$

The similarity measure directly participated in the parameter iteration through the loss function shown in Equation (3), which not only exceeded the traditional limitation of only relying on the measurement direction, but also ensured that the measurement of each layer was dynamically updated with the training process, effectively overcoming the limitations of static methods such as cosine similarity that could not be self-optimized.

(2) Sampling method

Since fraud is often accompanied by complex and variable impersonation methods, it is crucial to effectively identify and exclude these impersonation nodes. The limitations of existing sampling strategies are prominent, such as relying on human intervention with preset thresholds, which not only increases the dependence on developer experience and subjective judgment, but also introduces unnecessary randomness, which reduces the universality and stability of the method. Therefore, it is crucial to develop a strategy that can autonomously learn and dynamically optimize the sampling scale of neighbor nodes, aiming at realizing the intelligent and adaptive adjustment of the sampling process. The following formula is an attempt to solve this problem by dynamically calculating the number of nodes sampled.

$$Spl_{v,r}^{(l)} = p_r^{(l)} \cdot V_v^r \quad (4)$$

Specifically, $Spl_{v,r}^{(l)}$ is defined as the sampling number of node v under relation r in the LTH layer, $p_r^{(l)}$ is the filtering threshold set for relation r at this level, and $V_v^r$ represents the total number of neighbors of node v under relation r. After determining the number of sampled neighbors for each node, the sampling process is performed based on the similarity between them, and the

sampling formula is as follows:

$$C_{v,r}^{(l)} = Choose_{spl_{v,r}^{(l)}}\left(Sort(D^l(v,u))\right) \quad (5)$$

The Sort function is responsible for sorting the neighbors based on the calculated similarity, while $Choose_{spl_{v,r}^{(l)}}$ selects the neighbors with the first $spl_{v,r}^{(l)}$ name, and these nodes form $C_{v,r}^{(l)}$. That is, the set of neighbors of layer l node v after similarity filtering under relation r. This sampling strategy based on similarity effectively filters a large number of potential noisy neighbor data, and significantly enhances the ability to identify and exclude the masquerading behavior of fraudsters.

2. Core node reinforcers

(1) Weighted aggregation of central nodes

After completing the neighbor noise cancellation, the immediate next step is to update the information of the central node, which is achieved through the message passing process. The classical message passing mechanism can be summarized as follows:

$$H_{v,r}^{(l)} = ReLU\left(AGG_r^{(l)}\left(\{H_{v,v'}^{l-1}, v' \in N_r(v), v \in V\}\right)\right) \quad (6)$$

Here, $N_r(v)$ identifies the set of all neighbor nodes of node v under relation r, while $AGG_r^{(l)}$ represents the average aggregation operation of graph neural network performed on relation r at layer l.

Although the above message passing process borrows the mechanism of graph neural network to update the state of the central node, it still faces a challenge: even after neighbor noise filtering, a node with a large number of neighbors may still lead to residual noise information. It is difficult to completely eliminate all interference by relying on the noise filter alone, especially when the number of neighbors is large. In fact, the advantage of graph neural network in fraud detection is that it can effectively use the key neighbor information, but too much neighbor data may have a negative effect, as shown in Equation (7).

Therefore, it is necessary to introduce a weighting strategy for the central node in the aggregation step, that is, to increase the weight of the central node itself, which aims to ensure that the integrity of the central node features is maintained to the greatest extent while integrating neighbor information. This strategy aims to balance the relationship between the influence of global neighbors and maintaining the uniqueness of individual features, so as to improve the anti-noise ability and detection accuracy of the model.

$$h_r^{(l+1)} = \sigma\left(\widetilde{D}^{-\frac{1}{2}}\widetilde{A}\widetilde{D}^{-\frac{1}{2}}h_r^{(l)}W^{(l)}\right) \quad (7)$$

Here, $\widetilde{D}ii$ represents the degree of each node, and $\widetilde{A} = A + I$ is the adjacency matrix of the self-connected graph formed by the original adjacency matrix A plus the identity matrix I, which aims to consider the contribution of the node itself. $W^{(l)}$ for the training parameters, $h_r^{(l)} \in R^{N \times D}$ said in the activation of relation R first l layer node embedded vector, initialization time $h_r^{(0)} = X$.

Although the feature of the center node may be diluted due to too many neighbor nodes in the propagation process of Graph Convolutional Network (GCN), which has been alleviated by the above neighbor noise filtering step, it is particularly important to keep the feature of the center node fresh in the financial fraud detection scenario. Inspired by GCN, the consideration of the central node can be strengthened by increasing the weight of the self-loop edge, and the specific implementation is reflected in Equation (8). This strategy aims to balance the global information aggregation and protect the uniqueness of individual features, so as to optimize the performance of the model in complex fraud pattern recognition.

$$h_r^{(l+1)} = \sigma\left(\widetilde{M}^{-\frac{1}{2}}\widetilde{V}M^{-\frac{1}{2}}h_r^{(l)}W^{(l)}\right) \quad (8)$$

Where $h_r^{(l+1)}$ represents the updated embedding representation of each node at the l+1 layer under relation r, $\widetilde{V} = A + I \cdot (1 + p_r^{(l)})$ is the adjusted adjacency matrix, and the filtering threshold $p_r^{(l)}$ is introduced as the weight of the central node. This design aims to not only maintain the integrity of the characteristics of the central node, but also efficiently integrate the information from the neighbor nodes, so as to strengthen the characteristics of the central node without ignoring the surrounding environment information.

(2) Reinforcement learning module

Many studies adopt the average aggregation strategy in the aggregation step, which is simple but not efficient. Although the neighborhood information of a fraud node can reveal fraud features to some extent, its addition may inadvertently weaken the unique information of the central node. Therefore, it is particularly critical to build an update mechanism that can balance the information weight of the central node and its neighbor nodes. However, since $p_r^{(l)}$ is essentially a probability value and lacks derivability, it is impossible to directly use gradient descent method for parameter optimization. Therefore, when measuring the distance, it is particularly important to take into account the average distance between a fraudulent node and its neighbors, which can help to evaluate the relationship between nodes more comprehensively. The average distance is quantified as follows:

$$\overline{D}_r^{(l,e)} = \frac{\sum_{v \in V_{pos}} D_r^{(l)}(v,v')^{(e)}}{|V_{train}|}, v' \in C_{v,r}^{(l)}, r \in \{1, \dots, R\} \quad (9)$$

Here, $V_{pos}$ indicates the set of fraud nodes identified in the training phase, v' represents the nodes adjacent to the fraud node v, and $\overline{D}_r^{(l,e)}$ represents the average distance between the fraud node in level l and its corresponding neighbor nodes under the relationship r. Reinforcement learning (RL) mechanisms assign rewards or penalties based on this change in average distance: if the average distance of the current iteration has decreased compared to the previous one, a positive incentive is given; Conversely, if the distance increases, penalties will be imposed. Specific to the action policy, the decision logic of RL can be formulated as the following functional form:

$$f(p_r^{(l)})\begin{cases}+\tau, & \overline{D}_r^{(l,e-1)} - \overline{D}_r^{(l,e)} \geq 0 \\ -\tau, & \overline{D}_r^{(l,e-1)} - \overline{D}_r^{(l,e)} < 0\end{cases} \quad (10)$$

Here, τ represents the step size to update the threshold in reinforcement learning. The iteration process stops when a predetermined termination condition is reached, that is, either a predetermined number of iterations is reached or convergence is observed within 10 consecutive epochs, indicating that the optimal threshold has been found. Once the reinforcement learning module completes its task, the obtained $p_r^{(l)}$ value is not only used as the final neighbor filtering threshold, but also determines the weighting coefficient of the central node. The termination logic of reinforcement learning module can be formalized as follows.

$$\left|\sum_{e-10}^{e} f(p_r^{(l)})\right| \leq 2\tau, where\, e \geq 10\, and\, e \leq E \quad (11)$$

Here, e marks the current iteration round, while E is defined as the total iteration number bound set for the whole experiment.

3. Relationship aggregators

After enhancing the central node features, the next step is to perform aggregation across relations. This involves integrating the embedding $h_{v,r}^{(l)}$ of each node under a specific relation r with the embedding $h_v^{(l-1)}$ obtained by the node in the previous layer after the cross-relation aggregation to generate a comprehensive final embedding representation. See Equation (12) for the specific mathematical expression.

$$h_v^{(l)} = ReLU\left(W^{(l)}\left(h_v^{(l-1)} \oplus \{h_{v,r}^{(l)}\}|_{r=1}^R\right)\right) \quad (12)$$

Here, the symbol ⊕ refers to the concatenation operation, $h_{v,r}^{(l)}$ represents the embedding vector of node v at layer l under relation r, and $h_v^{(l-1)}$ represents the final embedding vector of node v at layer l-1 after aggregation.

For any node v, its final embedding in the model is represented as the output $h_v^{(L)}$ obtained by the GNN at the topmost L level. In order to evaluate the performance of the model and guide the optimization process, the cross-entropy loss function is selected as the main loss metric.

$$\mathcal{L}_{GNN} = -\sum_{v \in V} y_v log\left(MLP(h_v^{(L)})\right) + (1-y_v)log\left(1 - \left(MLP(h_v^{(L)})\right)\right) \quad (13)$$

B. *CNN layer*

The output produced by the GNN layer is used as the input data of the subsequent CNN layer. In the CNN layer, the convolution process can be summarized by Eq. (14), which generates the local feature vector ci, which involves the convolution kernel $w_j$, the bias term b, the convolution kernel size k, and the application of the activation function f.

$$c_i = f\left(\sum_{i=i-k}^{i+k} W_j x_j + b\right) \quad (14)$$

The description of the pooling step can be simplified by using equation (15), where pi refers to the pooling result, that is, the fixed-length feature vector, and k is the size of the pooling region. Then, the fully connected layer integrates all the local feature vectors $p_1, p_2, \ldots, p_n$ becomes an integrated global feature vector. This global feature vector is subsequently used as input information to the LSTM unit.

$$p_i = \max_{j=j-k}^{i+k} c_j \quad (15)$$

## C. LSTM layer

CNN layer output values are transformed by tanh function to generate vector set $X = \{x_1, x_2, \ldots, x_n\} \in \mathbb{R}^{n \times n}$, which is directly used as the input of the LSTM layer.

The vectors are fed into the bidirectional LSTM model and processed independently by the hidden layers in both directions. This includes forward and reverse hidden layers, which generate outputs at each time step. Specifically, the forward hidden state at the NTH time step is denoted as $\vec{h}_n$, while the reverse hidden state is denoted as $\overleftarrow{h_n}$, which is calculated according to equations (16) and (17).

$$\vec{h}_n = g(W_a^f x_i + W_{n-1}^f \overrightarrow{h_{n-1}}) \quad (16)$$
$$\overleftarrow{h_n} = g(W_a^b x_i + W_{n-1}^b \overleftarrow{h_{n-1}}) \quad (17)$$

In the formulation, $W_a^f$ represents the weight matrix from the input $x_i$ to the feedforward hidden layer, and $W_{n-1}^f$ is the weight matrix corresponding to the feedforward hidden layer output $\overrightarrow{h_{n-1}}$ at the previous time step n-1. Similarly, $W_a^b$ is the weight matrix of the input xi matched to the backward hidden layer, and $W_{n-1}^b$ involves the weight matrix of the backward hidden layer output at time n-1 $\overleftarrow{h_{n-1}}$.

Next, by merging the forward hidden state $\vec{h}_n$ at the current time step n with the backward hidden state $\overleftarrow{h_n}$, we obtain the integrated hidden layer output $h_n$ at that time step, which follows Equation (18).

$$h_n = \alpha(W^f \vec{h}_n + W^b \overleftarrow{h_{n-1}}) \quad (18)$$

In the formula, $W_f$ represents the weight matrix associated with the forward hidden layer output, while $W_b$ is the weight matrix corresponding to the backward hidden layer output; $\alpha$ uses a sigmoid activation function to introduce nonlinearity and control the degree of fusion of different hidden state information.

## IV. APPLICATION OF MODEL

### A. Dataset and evaluation metrics

1. Data sets

In order to test the fraud detection performance of the proposed GNN-CL model, the Yelp public data set is selected as the platform. Yelp data sets are widely used in business rating, sentiment analysis, personalized recommendation and regional search research. With their massive user feedback information on various types of local businesses, Yelp has become a treasure house in the fields of social network analysis, natural language processing and machine learning. The dataset is managed using the Linked Data methodology, which integrates multiple data formats, a fundamental component in academic research[29]. This systematic approach enables the cross-referencing of data, enhancing interoperability among various datasets. This capability is especially beneficial in fields such as machine learning and artificial intelligence, where the quality of data is critical for the effective training of models and the attainment of accurate results. The unique ternary structure of this dataset covers three core relationships of Yelp review spam datasets: repeated reviews by users (R-U-R), reviews with the same rating of the same item (R-T-R), and reviews with the same month of the same item (R-S-R), which deeply describes user behavior, review patterns, and time characteristics.

TABLE I. THE DATASET

| Dataset | Nodes | Edge | Relations |
|---|---|---|---|
| Yelp | 45954 | 49315 | R-U-R |
| | | 573616 | R-T-R |
| | | 3402743 | R-S-R |
| | | 3846979 | ALL |

2. Evaluation metrics

The evaluation metrics of classification models are important tools to measure the prediction performance of the model. They are used to judge the accuracy of the model when classifying unknown data.

a. Accuracy: This fundamental metric represents the ratio of correctly classified samples by the model to the overall sample count, reflecting its basic performance.

$$\text{Accuracy} = \frac{TP + TN}{TP + TN + FP + FN} \quad (19)$$

Where *TP* is a True Positive, *TN* is a True Negative, *FP* is a False Positive, and *FN* is a False Negative.

b. Precision: A measure of the fraction of examples that the model predicts to be positive are actually positive.

$$\text{Precision} = \frac{TP}{TP + FP} \quad (20)$$

c. Recall: Measure the proportion of all true positive examples found by the model, that is, the proportion of all actual positive examples identified by the model.

$$\text{Recall} = \frac{TP}{TP + FN} \quad (21)$$

d. F Score: A single measure of precision and recall that is the harmonic mean of them.

$$F = \frac{2 \times \text{Precision} \times \text{Recall}}{\text{Precision} + \text{Recall}} \quad (22)$$

In the experiments, we choose the well-known GCN algorithm as the basic reference, and compare the performance with the current state-of-art PC-GNN and CARE-GNN algorithms, so as to verify the superiority and practicability of our proposed algorithm in solving the fraud detection task based on graph structure.

### B. Experimental environment and parameters

1. Experimental environment

The implementation of GNN-CL is based on Pytorch version 1.10.2, and the dataset is evenly split with 40% of the training data and 60% of the test data. The environment is configured to run Python 3.8.5 on Ubuntu 18.04.1 with a GeForce RTX 2080 Ti GPU, 16GB RAM, and Intel Xeon Silver 4214 processor (2.20GHz). For PC-GNN and CARE-GNN, we directly adopt the code of the original author to conduct experiments. However, the experiment of GCN is implemented through the DGL library.

2. Experimental parameters

In view of the large number of nodes in the dataset, we adopt the mini-batch training strategy to efficiently train GNN-CL and various baseline models. We use Adam optimizer for parameter optimization, and the initial learning rate is set to 0.01. In the configuration of GNN-CL, the hyperparameters are set as E=50, L=1, $\lambda$ =2, and the step size of reinforcement learning $\tau$ =0.02. For Yelp dataset, the batch size is 1024 and the learning rate is 0.01. GCN, PC-GNN and CARE-GNN are also trained by Adam optimizer, and the configuration of iteration number, network layer number and batch size is consistent with GNN-CL on Yelp dataset to ensure the fairness and comparability of experiments. In particular, GNN-CL adopts a fixed weight threshold of 0.5 as a preset parameter to simplify hyperparameter tuning during model training.

### C. Experimental results and analysis

1. Performance comparison

Table 2 summarizes the performance comparison between our proposed GNN-CL and other basic algorithms on Yelp dataset, which clearly shows that GNN-CL has an advantage in most evaluation metrics. By integrating reinforcement learning into the feature enhancement process of the center node, the model effectively dealt with the challenge of the feature dilution of the center node when the graph neural network aggregated neighbor information. In contrast, state-of-the-art fraud detection algorithms CARE-GNN and PC-

GNN focus on solving the difficult problem of feature disguising and label imbalance.

The experimental results verify that our GNN-CL model outperforms CARE-GNN and PC-GNN in performance. Traditional graph neural network models, such as GCN, process node features according to the original adjacency matrix, which may contain noise and be non-optimized, further highlighting the advantages of GNN-CL innovative method.

TABLE II. FRAUD DETECTION PERFORMANCE OF EACH MODEL ON YELP DATASET

| Models | Precision Ratio | Recall Rate | F | Accuracy | AUC |
|---|---|---|---|---|---|
| GCN | 0.5994 | 0.6437 | 0.6208 | 0.7325 | 0.6849 |
| PC-GNN | 0.7937 | 0.8221 | 0.8077 | 0.8233 | 0.7595 |
| CARE-GNN | 0.8619 | 0.8311 | 0.8462 | 0.8559 | 0.7904 |
| GNN-CL | 0.8912 | 0.8878 | 0.8895 | 0.8969 | 0.8136 |

In the evaluation of Yelp dataset, GNN-CL shows significant advantages, surpassing the current top algorithms in all performance indicators, especially improving the F-score by 2.7 percentage points, and also surpassing the best existing algorithms in terms of Precision and Accuracy.

In order to verify the effectiveness of the proposed central node enhancement module, this study selected the Yelp data set for experiments, and set up the control group: one was the variant model GNN-CL-Y that removed the module, and the other was the model GNN-CL-W that adopted the fixed weight strategy. The experimental results are shown in Figure 2.

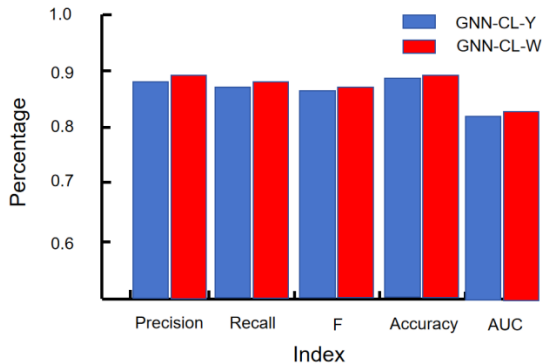

Figure 2. Impact of the central node enhancer on performance

The experimental results show that GNN-CL-W outperforms GNN-CL-Y in all evaluation indicators, which is specifically reflected in the AUC increased by 0.5 percentage points, the F1 score increased by 0.4%, the Recall increased by 0.3%, and the Precision and Accuracy were also improved. This strongly proves that the introduced central node enhancement module can effectively enhance the fraud detection performance of the model.

The key of graph convolution is to use the information of neighbor nodes for feature aggregation to enhance the classification performance. However, after multiple iterations, it is easy to cause the problem of node feature homogeneity and over-smoothing, that is, the difference between nodes decreases, which affects the discrimination ability. Therefore, the strategy of assigning differentiated weights to central nodes, as shown in this study, is an effective means to mitigate this problem and improve model performance.

2. Parameter sensitivity experiments

To evaluate the robustness of GNN-CL, we conduct experiments on the Yelp dataset, investigate its performance under varying parameter configurations, and conduct an exhaustive parameter sensitivity analysis. The experiments cover a variety of parameter Settings and show the corresponding sensitivity results. We present an in-depth comparison focusing on three key fraud detection performance metrics: AUC, F-score, and Recall[30].

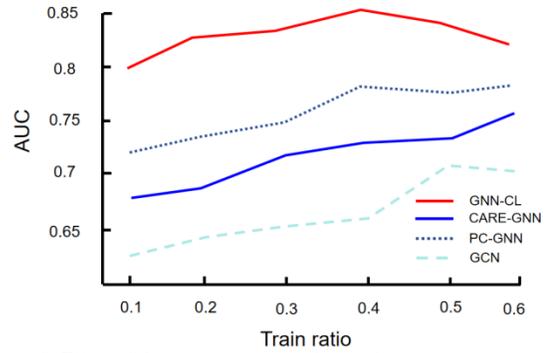

Figure 3. Effect of the training ratio on the results

Figure 3 shows the fluctuation of the AUC metric as the ratio of training data changes. It can be seen that no matter how the proportion of training set is adjusted, GNN-CL always maintains its superiority over the current optimal algorithm, and reaches its best performance when the training proportion is 40%.

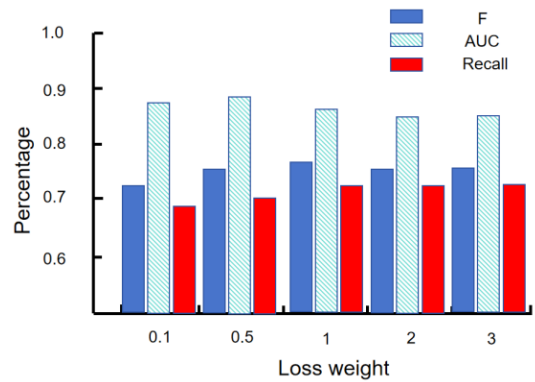

Figure 4. Effect of the loss weight on the results

Figure 4 shows the effect of varying the weight λ of the loss function on the performance of the model, which reveals that when λ is set to 0.5, the model has the best AUC performance. On the other hand, F-score and recall peak when λ is equal to 1.

V. CONCLUSIONS

This paper introduces an innovative fraud detection model, GNN-CL, which implements intelligent weighted aggregation of central nodes with the help of reinforcement learning. The core of the GNN-CL model consists of three layers of GNN, CNN, and LSTM, and is designed around three key modules: neighborhood noise purifier, core node intensifier, and relationship summarizer. Firstly, Multi-layer Perceptron (MLP) was used to predict node attributes and evaluate the similarity between nodes, and then the neighbors with low similarity to the central node were filtered out in the phase to reduce the interference of feature camouflage. Secondly, faced with the problem of weakening central features in aggregation, the central node reinforcement module used reinforcement learning to dynamically adjust the weight of central nodes. In this process, rewards and punishments are imposed according to the change of the distance between the fraud node and its neighbors (punishment is punishment when the distance increases, and reward is reward when the distance decreases) to ensure that the model gradually learns the strategy to optimize the weight of the central node. Finally, the node representations under each relation are summed up by the relation integrator to form a comprehensive feature representation for each node. Experiments on Yelp public datasets verify the effectiveness and advancement of ER-GNN in fraud detection tasks.